# Design and Analysis of Cold Gas Thruster to De-Orbit the PSLV Debris

**Roshan Sah, Raunak Srivastava, Kaushik Das**
Tata Consultancy Services (TCS)-Research
Bangalore, India
sah.roshan@tcs.com

## ABSTRACT

Today's world of space's primary concern is the uncontrolled growth of space debris and its probability of collision with spacecraft, particularly in the low earth orbit (LEO) regions. To prevent space debris growth, measures need to be taken so that there will be a debris depletion to conserve the LEO space environment and ease the risk of collision. This concern has developed the concept of Active Debris Removal (ADR). ADR method is an effective means for the removal of space debris using an active system. One of the ADR methods that have been continuously developed over a couple of decades is Space robotics whose function is to chase, capture and de-orbit the space junk. Our designed debris chaser satellite with dual robotic manipulators is made of a commercially available product of the market with an RCS thruster for the operation of orbit transfer, rendezvous, and close proximity operation and extra miniaturized cold gas thrusters for de-orbiting the polar orbit debris. Once reaching nearby to the PSLV debris, the robotic manipulators will be used to capture and cease the motion of the PSLV debris, and an extra cold gas thruster will be attached rigidly to the PSLV body using manipulators. Once, the attachment procedure is done, then de-orbiting to the altitude of 250 km will be done and the debris chase satellite will go on a search for other nearby debris. This paper is aimed to design an optimized micro-propulsion system, Cold Gas Thruster, to deorbit the PSLV debris from 668km to 250 km height after capturing process. The propulsion system mainly consists of a storage tank, pipes, control valves, and a convergent-divergent nozzle. The paper gives an idea of the design of each component based on a continuous iterative process until the design thrust requirements are met. All the components are designed in the CATIA V5, and the structural analysis is done in the ANSYS tool for each component where our cylinder tank can withstand the high hoop stress generated on its wall of it. And flow analysis is done by using the K-ε turbulence model for the CD nozzle, which provides the required thrust to de-orbit PSLV from a higher orbit to a lower orbit, after which the air drag will be enough to bring back to earth's atmosphere and burn it. Hohmann's orbit transfer method has been used to de-orbit the PSLV space debris, and it has been simulated by STK tools. And the result shows that our optimized designed thruster generates enough thrust to de-orbit the PSLV debris to a very low orbit.

## INTRODUCTION

After the end of their life, the post-mission removal of satellite structures has increased immense importance in future missions and keeps functional satellites in favorable orbit conditions. However, it becomes important to certify that de-orbiting operations take place in less time to limit debris collision probability. According to the ESA guidelines, it is mandatory to de-orbit the satellites within 25 years of the end of life.

In the future, it becomes important to preserve the LEO region environment with minimized risk [1]. Most of the articles show that spacecraft's uncontrolled space debris collisions had increased continuously [2]. Rex et al. studies show that the debris particle will grow with a 5% growth rate per year if the possible mitigating measures are not taken [3]. Anselmo et al. paper suggests that for the long-term evolution of debris population collision risk can be reduced by de-orbiting of upper rocket stages in LEO regions and by explosion avoidance strategies [4].

To prevent debris growth, it would be sufficient to remove 5-10 large pieces of the debris object per year [5].

There are two types of measures are defined to reduce the debris growth in an orbit, and they are:

1. Debris removal

2. Debris avoidance.

In debris removal, the space junks are removed by using a specific spacecraft or active propulsion system to the graveyard orbit which lies above the LEO region, or moving it to the very low earth orbit. And in case of debris avoidance, the operation spacecraft or satellite uses rendezvous maneuver by employing its propulsion system [6] to avoid the incoming space junks coming in their orbital path. Ornes suggests that those debris removals are one of the effective solutions but for small size junk, the removal method will be ineffective [7].

The Active Debris Removal (ADR) concept has continuously evolved for the debris removal from the LEO region. One of the evolving ADR methods is Space Robotics which has continuously matured over a couple of decades. In space robotics, the satellite system consists of a robotic arm mounted over it which can be called a Debris Chaser Satellite. The main function of the debris chaser satellite is to chase, capture and de-orbit the space junks to a very low earth orbit.

Our designed debris chaser satellite is made of dual robotic manipulators [8] [9], and the satellite subsystem is made commercially available in the market [10] [11]. The chaser satellite dimension is of 12U CubeSat standard whose main function is to chase, capture by the means of RCS thruster, and de-orbit by extra miniaturized cold gas thrusters for the PSLV debris removal from the polar orbit. The schematic diagram of our debris chaser satellite (without gripper) is shown in figure 1.

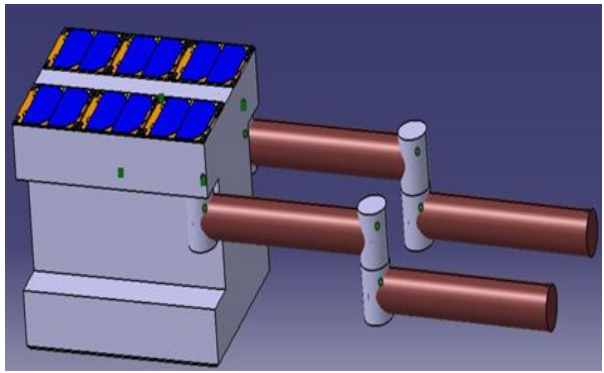

**Figure 1: Debris Chaser Satellite with robotic manipulators [8][20].**

This paper presents the design and analysis of the miniaturized cold gas thruster which will be used to remove the PSLV debris from its orbit to the very low earth orbit. This de-orbit operation can be done by Hohmann's orbit transfer method which involves using our designed thruster in a single direction. The PSLV debris removal can be done by the method of de-orbiting, which requires an active propulsion system. The propulsion system is a device used to propel the spacecraft from one place to another that uses propellant and oxidizer for its work. Burkhardt et. al. Studies show the advantage and drawbacks of different propulsion types on suitability for de-orbiting work and found that the cold gas propulsion system seems to be the right solution for de-orbiting work for all types of the satellite. However, the cold gas thruster is favorable for the small satellite's de-orbit function at low orbit with small delta-V requirements. He found that the spacecraft mass for small satellites with de-orbiting functions can be double [12].

This paper also presents the static and flow analysis of the propellant tank and the CD nozzle by using the

ANSYS software. Once a simulation is verified, the same cold gas thruster and its parameter will be used for the orbital simulation of PSLV debris for de-orbit operation by calculating the mass of propellant of a cold gas thruster.

**HOHMANN'S ORBIT TRANSFER**

Orbit transfer is an astrodynamics process of moving the satellite or space junk from one orbit to another orbit and Hohmann's transfer is most commonly used for orbit transfer. For our case, we will be using Hohmann's method for de-orbiting the PSLV debris. It is considered the most efficient and simplest method of transferring a satellite in co-planar orbits and the co-apsidal axis. It is a two-burn method for elliptical transfer between two coplanar circular orbits. The transfer method itself consists of an elliptical orbit with an apogee at the outer orbit and a perigee at the inner orbit. The schematic diagram of the typical Hohmann's orbit transfer method is shown in figure 2 with a single direction of net velocity after firing the thruster unit.

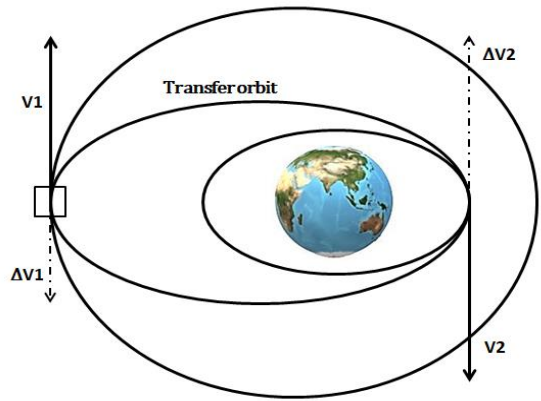

**Figure 2: Hohmann Transfer Orbit.**

Chemical propellants (solid or liquid) are used for Inter-planetary maneuvers like mars missions, etc. After many catastrophes caused by space debris, the major focus has been given to re-orbit and de-orbit a satellite. For the spacecraft orbiting at lower earth orbit, de-orbiting is preferred, and for the spacecraft orbiting at higher orbits, re-orbiting to Graveyard orbit is mostly preferred. The de-orbiting of a satellite is new, and rules for the small satellite's de-orbit have not been fully implemented. Various space organizations like ESA, NASA, and universities like Surrey space center, UTIAS SFL, DLR (German Aerospace Center) Braunschweig, JPL Caltech, University of Patras, etc. are working in de-orbit of a satellite or space junk. UTIAS SFL had successfully launched the CAN X series in which cold gas thruster is used for station keeping and currently working to develop cold gas thruster for de-orbiting operation.

## COLD GAS THRUSTER

The cold gas thruster is the most reliable thruster for the spacecraft and can be miniaturized to meet small satellites' design requirements. It is simple and cheap to design, has less power usage, is safe to handle, and operated in a continuous or pulse method that seems more promising than other thrusters [13]. It is the cheapest but limited-performing propulsion system in the aerospace industry [14]. Most aerospace companies are recently developing the micro-valves for cold gas thruster, which provides reliable thrust in the range of milli-Newton (mN)[15]. The cold gas thruster's main component is a pressurized tank, pipes, and solenoid valves to control propellant flow and nozzles, where the pressurized propellant is ejected [16]. Their detailed design regarding mass and volume has been investigated by Reichbach et al [17]. Moreover, their result suggests that the operating valve's response time is higher than the required time to achieve the required thrust. However, it is sufficient for orbit transfer and maneuvering applications. Multiple MEMS-based sensors developed by VACCO Company in the cold gas thruster field are the most versatile propulsion system available for CubeSat [18]. A cold gas thruster prototype was developed for small satellites by HSFL and it can be used for attitude control, de-orbiting, and station keeping, which generate a range of 0.1-10N thrust with a specific impulse of 75 s at LEO. It can generate about 50-70 m/s of delta-v requirement for de-orbits small satellites to lower orbit ranges [19].

A cold gas system consists of a pressurized tank containing propellant, flow control valves, solenoid valves, pressure gauges like pressure sensor 100, CD nozzle, and plumbing connecting them. The cold gas thruster works on the principle of conservation of energy. The gas's pressure energy is converted into kinetic energy by using a CD nozzle providing enough thrust to de-orbit the debris. The schematic diagram of the cold gas thruster is shown in figure 3.

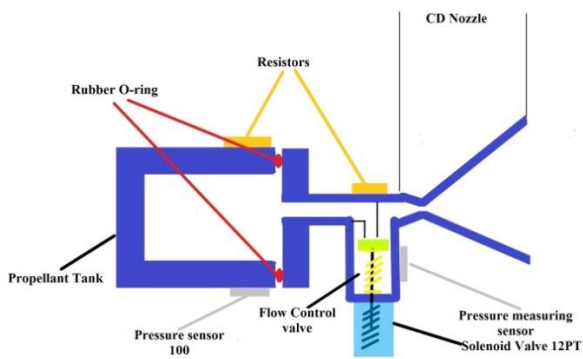

**Figure 3: Conceptual Design of Cold Gas Thruster.**

Any gas can be used as a propellant. However, the higher atomic mass gases are desirable as per Newton's Third Law, such as Helium, Nitrogen, Xenon, etc. To estimate the propellant's quantity, pressure, and temperature inside the tank, instrumentation devices are present. The valves control the propellant's release, and the nozzles accelerate the propellant to generate the desirable thrust. The cold gas thruster system does not contribute to any temperature rise nor generate any charge on the system hence known as Cold Gas.

### Component of Thruster

#### I. Storage Tank

Nitrogen gas is stored at very high pressure in the tank. Expansion of nitrogen takes place from higher pressure to space atmosphere where the pressure is negligible through the convergent-divergent nozzle. Tungsten matrix reinforced with boron fiber has a tensile strength of 3-4GPa and density of 2.23g/cc.

#### II. Pressure Regulating Valve

The valve is used to supply nitrogen at constant pressure to the CD nozzle. The pressure valve is controlled electronically based on time. The solenoid type of the pressure valves is commonly used in cold gas thruster.

#### III. CD Nozzle

The main function of the CD nozzle is to convert the pressure energy into kinetic energy. In our case, we use the De-Laval nozzle for supersonic flow which is also known as the CD nozzle.

#### IV. Choice of Propellant

The cold gas thruster uses different types of the propellants like Xenon, Helium, Nitrogen, Carbon dioxide, Sulphur hexafluoride, and Air. As nitrogen is easily available and its molecular mass is acceptable, it can be compressed at a higher temperature in a tank. Usually, nitrogen is non-corrosive at room temperature provided the amount of water vapor present in it is very less. While using nitrogen, the problem of corrosion doesn't arise. It can be stored at very high pressure and is very cheap.

## METHODOLOGY

This section started from the governing equation consisting of the Hohmann orbit transfer equation, chamber design, the nozzle design equation, and later design, meshing, and boundary conditions for structural and flow simulation. All the design is done based on the availability of the small satellites' volume and the de-orbiting application's thrust requirement.

### A. Governing Equations

The governing equations for modelling the cold gas thruster for our de-orbiting operation are shown below. It consists of equations of Hohmann orbit transfer, tank design, and nozzle design.

#### i. Hohmann transfer Governing Equation :-

$$V_{t1} = \sqrt{2[\left(\frac{\mu}{R_{orbit1}}\right) + \varepsilon_{transfer}]} \quad \dots\dots\dots (1)$$

$$\varepsilon_{transfer} = \frac{-\mu}{2a_{transfer}}$$

$$2\,a_{transfer} = \ R_{orbit1} + R_{orbit2}$$

$$V_{orbit1} = \sqrt{2[\left(\frac{\mu}{R_{orbit1}}\right) + \varepsilon_{orbit1}]} \quad \dots\dots\dots (2)$$

$$\varepsilon_{orbit1} = \frac{-\mu}{2R_{orbit1}}$$

$$\Delta V_1 = |V_{orbit1} - V_{transfer\ orbit}| \quad \dots\dots\dots (3)$$

$$V_{t2} = \sqrt{2[\left(\frac{\mu}{R_{orbit2}}\right) + \varepsilon_{transfer}]} \quad \dots\dots\dots (4)$$

$$V_{orbit2} = \sqrt{2[\left(\frac{\mu}{R_{orbit2}}\right) + \varepsilon_{orbit2}]} \quad \dots\dots\dots (5)$$

$$\varepsilon_{orbit2} = \frac{-\mu}{2R_{orbit2}}$$

$$\Delta V_2 = |V_{orbit2} - V_{transfer\ orbit}| \quad \dots\dots\dots (6)$$

$$\Delta V = \Delta V_1 + \ \Delta V_2 \quad \dots\dots\dots\dots (7)$$

$$TOF = \pi \sqrt{\frac{a_{transfer}^2}{\mu}} \quad \dots\dots\dots\dots (8)$$

#### ii. Propellant mass calculation

$$\Delta V = u_{eq} ln\frac{m_i}{m_f} \quad \dots\dots\dots\dots (9)$$

$$m_p = m_i - m_f$$

#### iii. Tank design equation

$$P_c = \frac{b}{r_i^2} - a \quad \dots\dots\dots\dots (10)$$

$$P_o = \frac{b}{r_e^2} - a \quad \dots\dots\dots\dots (11)$$

$$f = \frac{b}{r_i^2} - a \quad \dots\dots\dots\dots (12)$$

#### iv. Nozzle design equation

$$\frac{P_c}{P_e} = (1 + \frac{\gamma-1}{2}M_e^2)^{\frac{\gamma}{\gamma-1}} \quad \dots\dots\dots(13)$$

$$\frac{T_c}{T_e} = \ (1 + \frac{\gamma-1}{2}M_e^2)$$

$$\varepsilon = \frac{1}{M_e}\sqrt{[\frac{2}{\gamma+1}(1+\frac{\gamma-1}{2}M_e^2)]^{\frac{\gamma+1}{\gamma-1}}} \quad \dots\dots (14)$$

$$\varepsilon = (\frac{A_e}{A_t})$$

### B. Design

The design process always starts from the conceptual design, where the propellant volume's availability is considered to design our fully optimized thruster. The parameters like the volume of propellant, size of the tank, temperature, pressure inside the tank, dimension of CD nozzle, and the dimension of the pipes are all calculated iteratively by equations 1-14 until an optimized design of thruster is obtained. The thruster's preliminary design is done solely based on the delta-V requirement determined by the Hohmann orbit transfer equation 1-8. From the table-1, we can see that the delta-V is obtained at -223.5 m/s (negative sign represent the case of de-orbiting from a higher orbit to a lower orbit), and the de-orbiting timing is 46.8 min. And using the delta-V value in determining the propellant's mass required for designing the tank and the thruster CD nozzle. As for cold gas thruster, we will be using the nitrogen gas whose specific impulse is the 80s, and the calculated mass of nitrogen propellant required for de-orbiting operation is about 2.27*10$^{-4}$ m$^3$. The tank design is done considering the volume space of debris chaser satellites and the volume of the N$_2$ propellant to be stored in the tank. The maximum chamber pressure of 30 bar, the hoop stress of 4 bar, and the safety factor of 1.5 is fixed while designing the propellant tank. This input data can be incorporated in equations 10-12 to find the required internal diameter, length, and chamber thickness which were tabulated in table 2. It was found that the calculated internal diameter and length of the tank are 75 mm and 52 mm approximately.

**Table 1 Orbit transfer from 668km to 250 km parameter**

| Parameters | Calculated values | Unit |
|---|---|---|
| $R_{earth}$ | 6378.14 | Km |
| $R_{orbit1}$ | 7046.14 | Km |
| $R_{orbit2}$ | 6628.14 | Km |
| $a_{transfer}$ | 6837.14 | Km |
| $\varepsilon_{transfer}$ | -29.149 | $Km^2/s^2$ |
| $\varepsilon_{orbit1}$ | -28.284 | $Km^2/s^2$ |
| $\varepsilon_{orbit2}$ | -30.068 | $Km^2/s^2$ |
| $V_{t1}$ | 7.405 | Km/s |
| $V_{orbit1}$ | 7.521 | Km/s |
| $\Delta V1$ | -1.1589 | Km/s |
| $V_{t2}$ | 7.872 | Km/s |
| $V_{orbit2}$ | 7.754 | Km/s |
| $\Delta V2$ | -1.176 | Km/s |
| $\Delta V$ | -0.2335 | Km/s |
| TOF | 2813.15 | sec |

**Table 2 Tank Dimension's.**

| Parameters | Calculated Values | Unit |
|---|---|---|
| Inlet diameter | 75 | mm |
| length | 52 | mm |
| Thickness | 3 | mm |

**Table 3 Nozzle Dimensions**

| Parameters | Calculated Values | Unit |
|---|---|---|
| Throat Area | 19.64 | $mm^2$ |
| Exit Diameter | 6.96 | m |
| Inlet diameter | 10 | mm |
| Nozzle half angle | 5 | degree |
| Convergent length | 11.44 | mm |
| Divergent length | 9.47 | mm |

The delta-V and the calculated mass of propellant values further help determine the expansion ratio of the CD nozzle, which helps to determine the nozzle parameter based on the flow control valve arrangement, which is shown in table 3. From table 3, we can see that the inlet and exit diameter is about 10 mm, and 6.96 mm approximate. The nozzle throat diameter is calculated in such a way that, it will always remain at Mach number one. The basic modelling principle is followed while designing the CD nozzle where the convergent section remains in subsonic condition and the divergent section remains in a supersonic condition which will provide high exit velocity. Hence, the thrust at the exit will remain maximum. Based on the design calculation, we make our CAD design of the cold gas thruster, which consists of the propellant tank, valve arrangement, CD nozzle, and pressure measuring sensor.

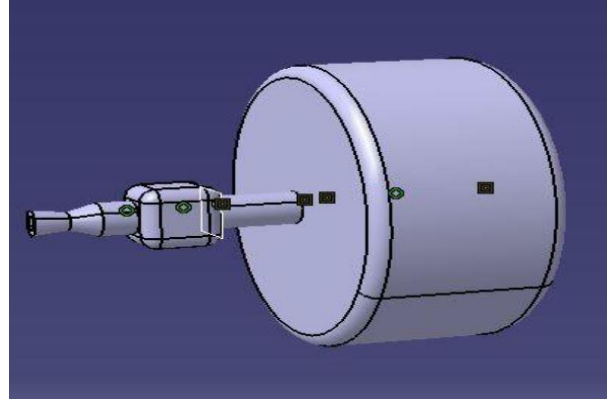

**Figure 4: CAD design of Cold Gas Thruster for De-orbiting Operation.**

The schematic CAD designs of the cold gas thruster for PSLV debris operation are shown in figure 4. This CAD file will be going for the ANSYS simulation which involves meshing, CFX, boundary condition insertion, and post-processing, which will help to validate our designed cold gas thruster.

### C. Meshing

The thruster 3D CAD design is subjected to meshing, where it divides the geometries into simple nodes and the element. As the mesh influences the convergence, accuracy, and simulation speed, the structured meshes were done for the chamber tank and the CD nozzle of the thruster. The structured mesh is generated by the ANSYS ICEM tool. The mesh was refined using O-grid to capture the boundary phenomena.

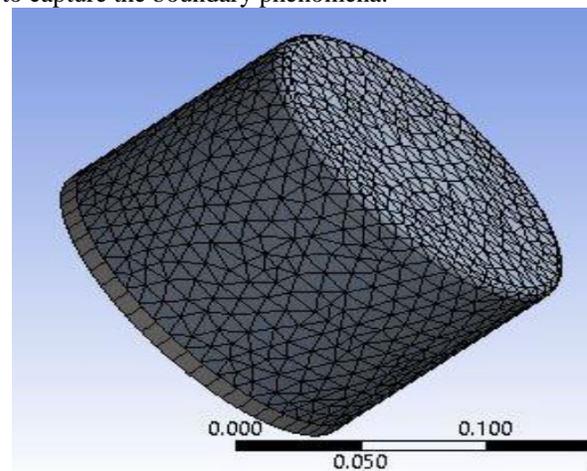

**Figure 5: Meshing of the Propellant tank.**

The skewness for the mesh was 0.7 to 1, where 0 is considered the worst and ideal. The generated meshes of the thruster tank undergo structural analysis by using the ANSYS Static structure tool. Moreover, the CD nozzle undergoes the flow simulation by the ANSYS CFX tool. The schematic diagram of a mesh of the thruster tank is shown in figure 5.

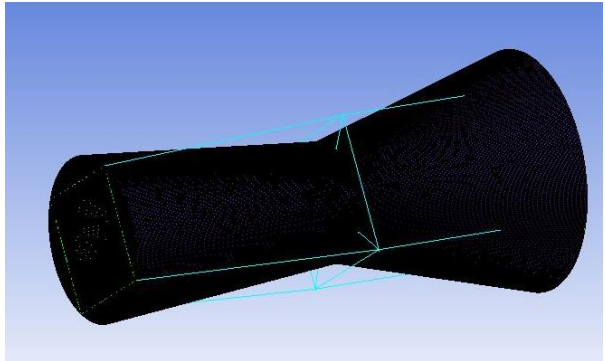

**Figure 6: Structure Mesh of CD nozzle by using O-grid.**

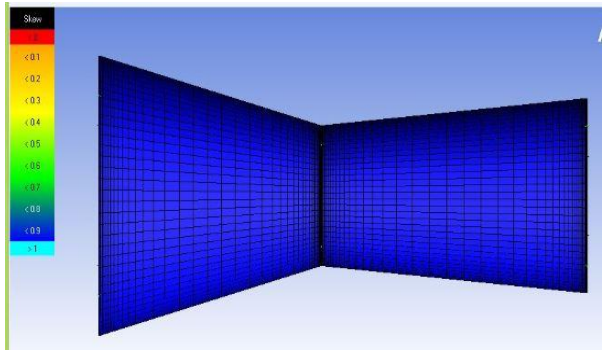

**Figure 7: Front view of refined mesh of CD nozzle.**

The schematic diagram of the structure mesh of the CD nozzle is shown in figure 6. Whereas, the CD nozzle is refined by using the O-grid operation in ANSYS ICEM to capture the flow properties with static wall conditions from inlet to outlet. The refined mesh of the CD nozzle by using an O-grid is shown in figure 7 which is the front view of it. The mesh description of the cd nozzle is listed in table 4.

#### D. Boundary Conditions

Once the meshing is done for the pressure tank and the CD nozzle, the mesh file is moved to the ANSYS static and CFX tool where the required boundary condition is applied for the simulation. The internal pressure and hoop stress of the tank are used as a boundary condition, which is taken as 30 bar and 4 bar with a safety factor of 1.5. These factors help us find the thickness of the chamber, which can be incorporated into structural analysis through ANSYS static Structure tools.

The density-based solver is used to simulate the CD nozzle with pressure inlet boundary condition applied at the inlet with an inlet pressure of 1 bar and temperature of 290K. The no-slip condition was imposed on the walls in the normal direction with an adiabatic heat transfer condition. Whereas pressure outlet boundary condition is used with a pressure of 0.1 bar.

To model the turbulence activity, the k-ε model is used as a turbulence model which solves 1 continuity, 3 momentum, and 2 turbulence kinetic and dissipation equations in the simulations.

**Table 4 Mesh description and BC.**

| Parameters | Calculated Values |
|---|---|
| No. of Nodes | 125989 |
| No. of Elements | 121385 |
| No. of hexahedrons | 121385 |
| No. of faces | 10958 |
| Reference pressure | 0.1 bar |
| Inlet pressure | 1 bar |
| Inlet temperature | 290 K |
| Turbulence model | K-ε model |

The mesh description and the boundary condition for the nozzle are tabulated in table 4.

### ANALYSIS

The values obtained by mathematical calculations are verified using MATLAB code. The essential parameters like exit velocity of air from the nozzle, maximum hoop stress in the tank, and maximum deformation of the tank under the influence of internal pressure are simulated using ANSYS.

#### 1. De-orbiting Trajectory

To simulate the de-orbit phenomenon by the Hohmann transfer method AGI STK tool is used. STK gives the orbital parameters latitude, longitude, and altitude during the de-orbit period. After every point of time, the position of the satellite is determined while de-orbiting the satellite from 668km to 250km. The main reason for using STK is the reliability of results used by NASA and other organizations for various space maneuvers simulations.

*2. Collision Avoidance*

The de-orbiting of space debris is a promising and elegant solution to the space debris removal problem. However, during de-orbiting, there is the probability of collision of de-orbiting space junk with the functional satellite, which again creates a severe problem. Several scenario simulations performed show that collision odds are very low, but a better future of de-orbiting collision avoidance study is essential. Active methods like thruster to avoid collision are not applicable for the small satellite due to size, mass, and other constraints. If the trajectory of the de-orbiting satellite and the orbit of the operational satellite intersect and the de-orbiting satellite and functional are present at that point at the same time, a collision occurs. After realizing this, an extensive study on the orbital dynamics and TLE was done. TLE is Two Line Element, which contains information about the revolving body like a satellite. From TLE, we can find the satellite's nearly exact position, and then the de-orbit maneuver is started such that the de-orbiting satellite. The TLE of the different satellites is read in the STK and the de-orbiting operation was executed neatly without any collision with any object presented at LEO orbit during transfer.

## RESULTS AND DISCUSSION

The static structural analysis of the propellant storage tank in the ANSYS workbench shows that the maximum equivalent Von-Mises stress is 1.12 Gpa, and the minimum value is obtained at 1.9 Mpa. It can be found that the maximum stress obtained is lower than the permissible safe with a load factor of 1.5. The schematic diagram of the Von-Mises stress distribution inside the thruster tank is shown in figure 8.

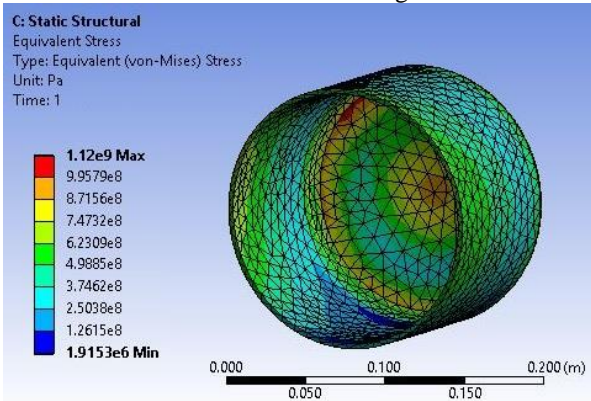

**Figure 8: Von-Mises Stress inside the propellant tank.**

From figure 8, we can see that maximum stress is near the tank's base periphery and the center of the tank. Similarly, the schematic diagram of the total deformation caused due to the internal pressure effect inside the propellant tank is shown in figure 9. From figure 9, we can see that the maximum total deformation of the cold gas thruster tank occurs at the center of the tank's base, which is nearly 2.319mm approximate, which is within our acceptance range.

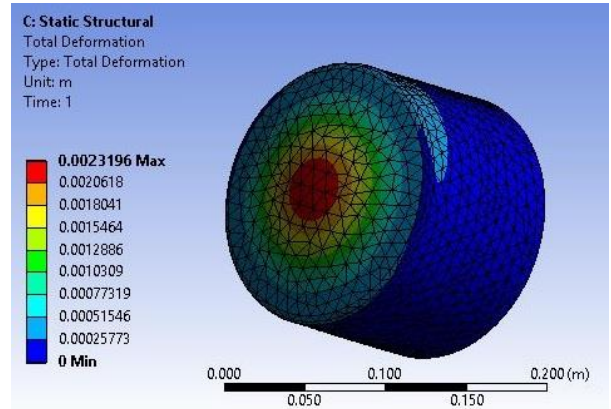

**Figure 9: Total deformation of pressurized propellant tank.**

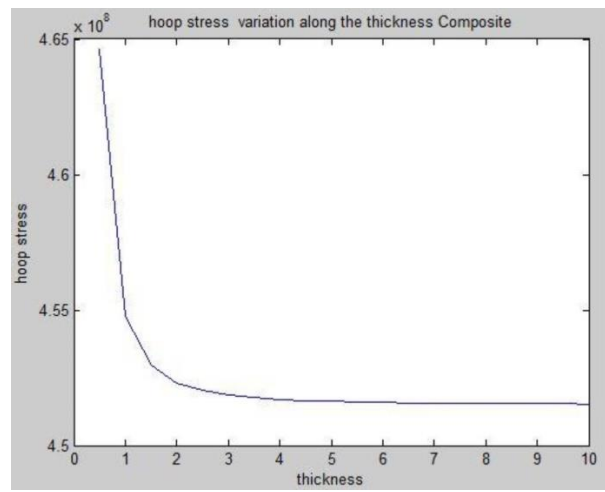

**Figure 10: Variation of hoop stress along thickness of tank.**

The graphical representation of hoop stress variation along the thickness of the storage tank is shown in figure-10. It can be found that the hoop stress varies exponentially from the internal thickness up to nearly 2 mm of thickness, and then the hoop stress is constant up to the external thickness section. Hence, it can be verified that the selected thickness for the thruster tank which is 3 mm approximate (shown in table 2), is within the permissible limit.

Once the propellant tank design is done and rectified by the static structural analysis, we can go for the simulation of the CD nozzle based on our design.

The result obtained from the ANSYS CFX for the CD nozzle is closely matched with the calculation done for the CD nozzle dimension.

The various results, like pressure, Mach number, and temperature effects, are obtained from the CFX post-processing data. The schematic contour diagram of the pressure distribution across the length of the nozzle is shown in figure 11.

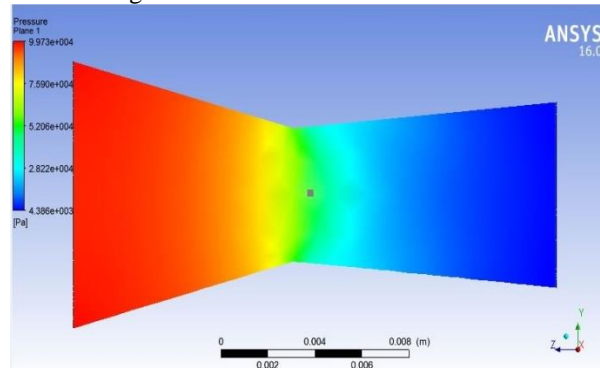

**Figure 11: Contour plot of pressure distribution across the length of the CD nozzle.**

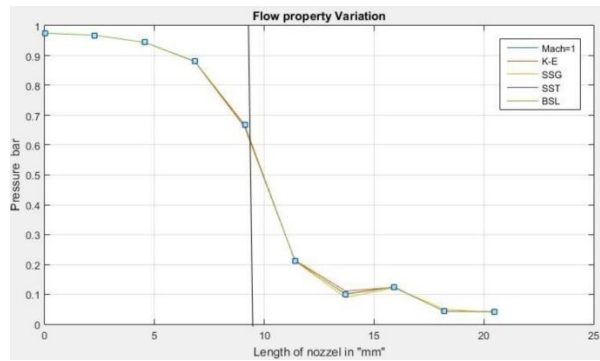

**Figure 12: Variation of pressure across the length of CD nozzle.**

From figure 11, we can see that the maximum pressure is 0.997bar, which is approximately the same as the initial input pressure of 1 bar, and the minimum pressure obtained is 0.044bar. From the figure11, we can see that pressure is continuously decreasing along with the nozzle's central length from input to output, which can be shown in the graphical diagram in figure 12. In figure 12, the effect is captured by considering the different turbulence models like K-ε, SSG Reynolds stress, Shear Stress Transport (SST), and the BSL Reynolds number. All these models follow the same pressure flow line with some of the line's deviation due to the lack of the incomplete flow modelling equations in some of the turbulence models. Furthermore, we can see that the throat where the Mach number is 1 has a pressure of 0.62bar.

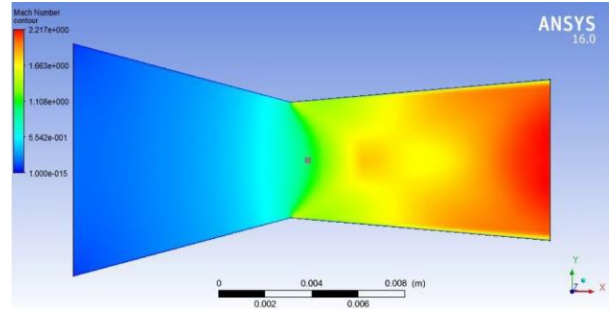

**Figure 13: Contour plot of Mach number distribution across the length of the CD nozzle.**

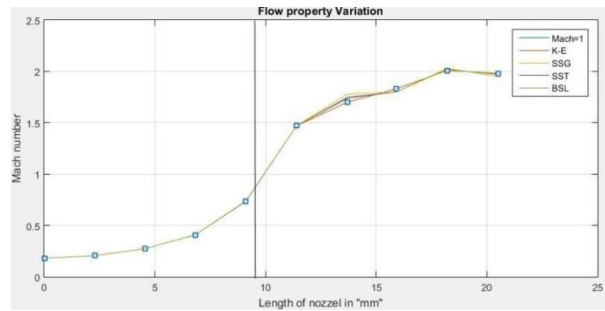

**Figure 14 : Variation of Mach number across the length of the CD nozzle.**

The schematic contour diagram of Mach number distribution across the length of the CD nozzle is shown in figure 13. From figure 13, we can see that the minimum Mach number, which is at the inlet, is less than 0.1 because we had applied pressure boundary conditions, and the maximum Mach number is 2.22 approximate, which is at at outlet. So, we can see that the Mach number is continuously increasing from the inlet to the throat section at the convergent section and throat section, Mach number reaches 1 (which is the chocked condition). After onward, the Mach number continuously increases to the divergent section's outlet due to the supersonic flow. Hence, for the same, the graphical representation of Mach number propagation at the central line along the length of the nozzle is shown in figure 14. Moreover, these flow properties are captured by the different turbulence models, as mentioned for the pressure effect. The slight variation can be seen in the graph due to the incomplete capturing of flow properties by some turbulence models.

Similarly, the schematic contour diagram of temperature distribution across the length of the nozzle is shown in figure 15. From figure 15, we can see that the minimum temperature, is 163.1K at the outlet section, and the maximum temperature is 289.9 K, which is the same as the inlet boundary condition. We can see from figure 15, that the temperature effect is continuously decreasing from the inlet to the outlet section. At the throat section, it is 258K. Hence, for the same, the graphical representation of temperature propagation at the central line along the length of the nozzle is shown in figure 16.

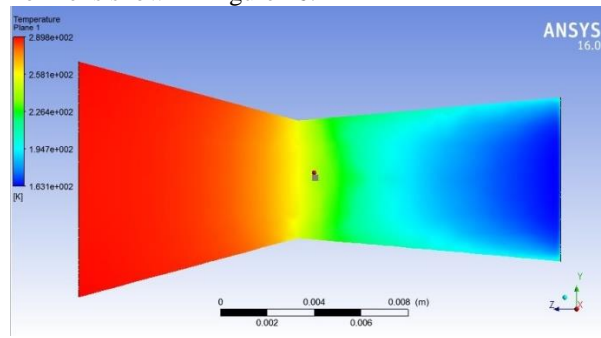

**Figure 15: Contour plot of Temperature distribution across the length of CD nozzle.**

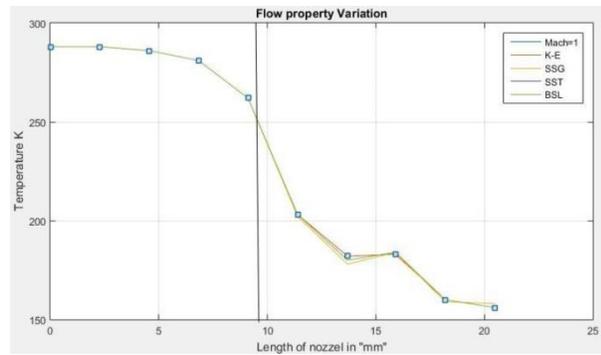

**Figure 16: Variation of Temperature across the length of CD nozzle.**

Furthermore, the temperature flow properties are captured by all turbulence models, as mentioned for the pressure effect. Moreover, the variation can be seen in the plot due to the incomplete capturing of flow properties by some other turbulence models in the simulation.

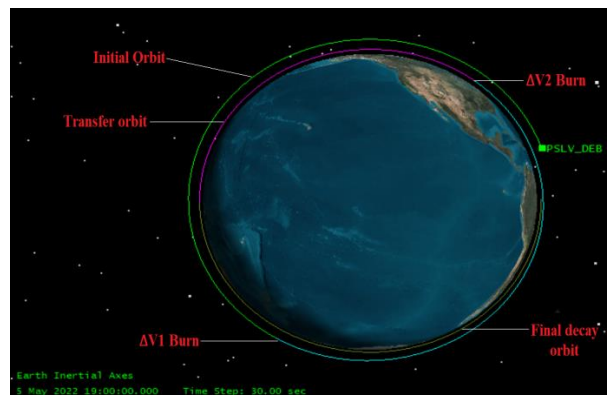

**Figure 17: De-orbiting of PSLV debris from its orbit to very low orbit.**

The orbital simulation [20] sketch of de-orbiting the PSLV debris is shown in figure 17 and the zoomed PSLV rocket staging body is shown in figure 18. The designed cold gas thruster and its parameter are used while doing the simulation. It was found that based on the calculated mass of propellant is enough to de-orbit the PSLV debris to a very low earth orbit i.e. 250 km.

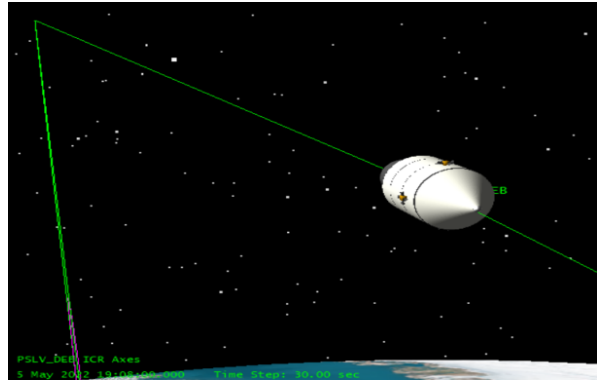

**Figure 18: Zoom view of PSLV staging rocket body for de-orbiting operation.**

From figure 17, we can see that the light green line as the initial PSLV orbit which is at an altitude of 668 km, light blue (ΔV1) and pink color (ΔV1) as the Hohmann orbit transfer, and dark green as the final PSLV decay orbit at 250 km.

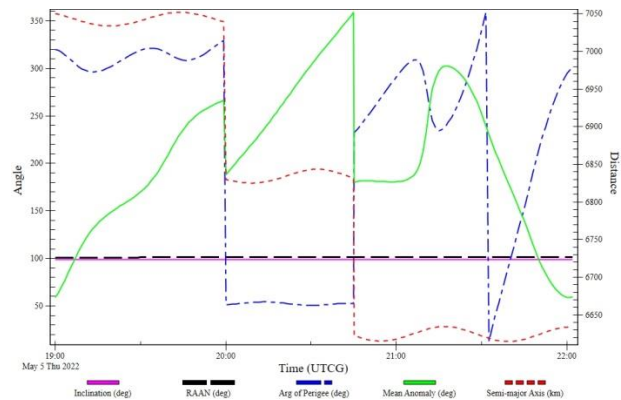

**Figure 19: Orbital elements variation of PSLV debris from 668 km to 250 km altitude by using Cold gas thruster.**

The variation of the orbital elements of PSLV debris from its orbit to very low orbit i.e. 250 km caused by cold gas thruster is shown in figure 19. From figure 19, we can see that the semi-major axis is continuously decreasing at ΔV1 and ΔV2 burn by Hohmann's orbit transfer method. The semi-major axis variation causes other orbital parameter changes along the simulation time except for inclination and RAAN.

After onward, the debris satellites are continuously interacting with the drag force due to the earth's atmosphere and causing the decay of the satellites into a fireball.

## CONCLUSION AND FUTURE WORKS

Deorbiting the spent satellite is an elegant and promising solution to the space debris problem. The PSLV debris is de-orbited by using the designed cold gas thruster which was allocated at the debris chaser satellite. Our debris chaser satellite is the dual robotic manipulator's satellite whose main function is to chase, capture and de-orbit the large space junk like PSLV rocket bodies, spent satellites, etc. Once the capturing operation is done by the debris chaser satellite by using robotics manipulator's, it uses an RCS thruster to stabilize the random PSLV debris then an extra miniaturized cold gas thruster will be attached to the debris object and it will go into de-orbiting operation. Whereas, the debris chaser satellites will go for other nearby debris locations for capturing operation.

The cold gas thruster is one of the simplest, most efficient, less expensive and most convenient propulsion systems that do not create a temperature gradient or ionize the metallic components. This paper demonstrates the assembled Cold Gas Thruster's optimized design and simulation and its application to de-orbit satellites from higher LEO to lower LEO to minimize space debris. It has also presented the thruster's structural and flow analysis, which was taking place from the tank to the nozzle section of the thruster. The structural and flow simulation shows promising results for developing the cold gas thruster for debris removal operations.

Our designed cold gas thruster has successfully de-orbited the PSLV debris from its orbit (i.e. 668 km) to 250 km of altitude by the calculated amount of propellant of a cold gas thruster. Once orbiting at 250 km, where the air drag will be enough to bring it back to the earth's atmosphere and burn it.

In future works, we are planning to fabricate the optimized cold gas thruster for de-orbiting and attitude control application for controlling the orientation of the satellites and removal of the space debris object.